\documentclass[a4paper]{article}
\usepackage{iwslt18,amssymb,amsmath,epsfig}
         \def\reg{{\rm\ooalign{\hfil
     \raise.07ex\hbox{\scriptsize R}\hfil\crcr\mathhexbox20D}}}

\usepackage{times}
\usepackage{latexsym}
\usepackage{makecell}
\usepackage{soul}
\usepackage{url}
\usepackage{color}
\usepackage{inconsolata}

\newcommand\rephrase[1]{{\emph{#1}}}
\newcommand\drop[1]{\textcolor{red}{#1}}
\newcommand\tense[1]{\ul{#1}}

\title{Controlling the Output Length of Neural Machine Translation}

 \makeatletter
 \def\name#1{\gdef\@name{#1\\}}
 \makeatother
 \name{{\em 
 Surafel Melaku Lakew$^{\star \dagger}$\hspace{0.2cm} 
 Mattia Di Gangi$^{\star \dagger}$\hspace{0.2cm} 
 Marcello Federico\thanks{Authors are in random order and contributed equally. 
  The first two authors carried out the work during their internship at Amazon. } }}

\address{$^{}$ Amazon AI - Palo Alto, USA\\$^{\star}$Fondazione Bruno Kessler, Trento - Italy\\$^{\dagger}$ University of Trento, Italy }

\begin{document}
\maketitle
\begin{abstract}
The recent advances introduced by neural machine translation (NMT) are rapidly expanding the application fields of machine translation, as well as reshaping the quality level to be targeted. In particular, if translations have to fit some given layout, quality should not only be measured in terms of adequacy and fluency, but also length.  Exemplary cases are the translation of document files, subtitles, and scripts for dubbing, where the output length should ideally be as close as possible to the length of the input text.
This paper addresses for the first time, to the best of our knowledge, the problem of controlling the output length in NMT.  
We investigate two methods for biasing the output length with a transformer architecture: i) conditioning the output to a given target-source length-ratio class 
and ii) enriching the transformer positional embedding with length information.
Our experiments show that both methods can induce the network to generate shorter translations, 
as well as acquiring interpretable linguistic skills.
\end{abstract}

\section{Introduction}\label{sec:intro}
The sequence to sequence \cite{bahdanau2014neural, sutskever2014sequence} approach to Neural Machine Translation (NMT) has shown to improve quality in various translation tasks \cite{bentivogli:CSL2018,hassan18,laubi18}. While translation quality is normally measured in terms of correct transfer of meaning and of fluency, there are several applications of NMT that would benefit from optimizing the output length, such as the translation of document elements that have to fit a given layout -- e.g. entries of tables or bullet points of a presentation -- or subtitles, which have to fit visual constraints and readability goals, as well as speech dubbing, for which the length of the translation should be as 
close as possible to the length of the original sentence.

Current NMT models do not model explicitly sentence lengths of input and output, and the decoding methods do not allow to specify desired number of tokens to be generated. Instead, they implicitly rely on the observed length of the training examples \cite{murray2018correcting, shi2016neural}.

Sequence-to-sequence models have been also applied to text summarization \cite{rush2015neural} to map the relevant information found in a long text into a limited-length summary. Such models have shown promising results by directly controlling the output length \cite{kikuchi2016controlling, fan2017controllable, liu2018controlling, takase2019}. However, differently from MT, text summarization (besides being a monolingual task) is characterized by target sentences that are always much shorter than the corresponding source sentences. While in MT, the distribution of the relative lengths of source and target depends on the two languages and can significantly vary from one sentence pair to another due to stylistic decisions of the translator and linguistic constraints (\textit{e.g.} idiomatic expressions).

% \begin{table*}[t]
% \small
% \begin{center}
% \begin{tabular}{p{0.5cm}|p{17cm}|}
% \hline
% SRC & {\footnotesize \tt It is actually the true integration of the man and the machine.}\\
% MT  & {\footnotesize \tt Es ist \underline{tats\"achlich} die \underline{wahre} Integration von Mensch und Maschin\red{e.}}\\
% MT* & {\footnotesize \tt Es ist die \underline{wirkliche} Integration von Mensch und Maschine.\blue{------}}\\
% \hline
% SRC & {\footnotesize \tt So we thought we would look at this challenge and create an exoskeleton that would help deal with this issue.}\\
% MT & {\footnotesize \tt \underline{Quindi abbiamo pensato} di guardare a questa sfida e creare un esoscheletro che potesse aiutare \underline{ad affrontare} \underline{ques\red{to}} \red{problema.}}\\
% MT* & {\footnotesize \tt \underline{Pensavamo} di guardare a questa sfida e creare un esoscheletro che potesse aiutare \underline{a risolvere} il problema.\blue{---}}\\
% \hline
% \end{tabular}
% \end{center}
% \caption{\label{Example} German and Italian human and machine translations (MT) are usually longer than their English source (SRC). We investigate enhanced NMT (MT*) that can also generate translations shorter than the source length. Text in red exceeds the length of the source, while underlined words point out the different translation strategy of the enhanced NMT model.}
% \end{table*}

\begin{figure*}[t]
    \centering
    \includegraphics[width=\textwidth]{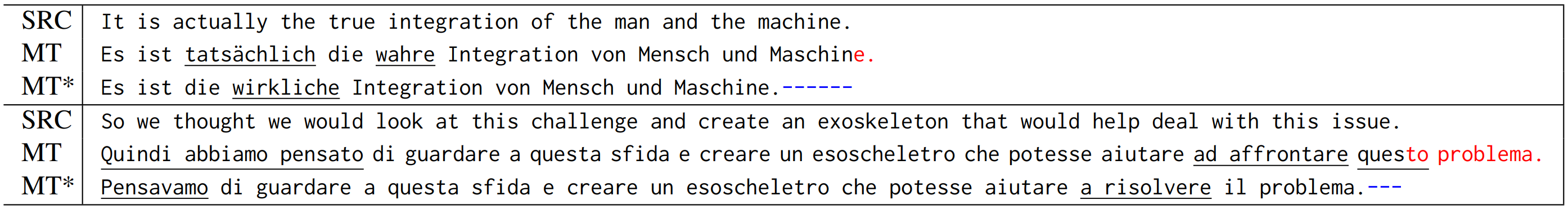}
    \caption{\label{Example} German and Italian human and machine translations (MT) are usually longer than their English source (SRC). We investigate enhanced NMT (MT*) that can also generate translations shorter than the source length. Text in red exceeds the length of the source, while underlined words point out the different translation strategy of the enhanced NMT model.}
\end{figure*}

In this work, we propose two approaches to control the output length of a transformer NMT model. In the first approach, we augment the source side with a token representing a specific length-ratio class, i.e. {\it short, normal, and long}, which at training time corresponds to the observed ratio and at inference time to the desired ratio.
In the second approach, inspired by  recent work in text summarization \cite{takase2019}, 
we enrich the position encoding used by the transformer model with information representing the position of words with respect to the end of the target string.

We investigate both methods, either in isolation or combined, on two translation directions (En-It and En-De) for which the length of the target is on average longer than the length of the source. Our ultimate goal is to generate translations whose length is not longer than that of the source string (see example in Table~\ref{Example}). While generating translations that are just a few words shorter might appear as a simple task, it 
actually implies good control of the target language. As the reported examples show, the network has to implicitly apply strategies such as choosing shorter rephrasing, avoiding redundant adverbs and adjectives, using different verb tenses, etc. 
We report MT performance results under two training data conditions, small and large, which show limited degradation in BLEU score and n-gram precision as we vary the target length ratio of our models. We also run a manual evaluation which shows for the En-It task a slight quality degradation in 
exchange of a statistically significant reduction in the average length ratio, from 1.05 to 1.01.

\begin{figure*}[t]
    \centering
    \includegraphics[width=0.8\textwidth]{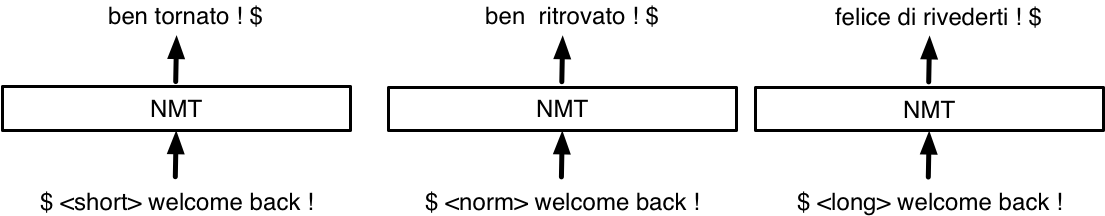}
    \caption{Training NMT with three length ratio classes permits to get outputs of different length at inference time.}
    \label{fig:token}
\end{figure*}

\section{Background}
Our proposal is based on the transformer architecture and a recently proposed extension of its positional encoding aimed to control the length of generated sentences in text summarization.

\subsection{Transformer}
Transformer \cite{vaswani2017attention} is a sequence-to-sequence architecture that processes sequences using only attention and feed forward layers. Its core component is the so-called multi-head attention, which computes attention \cite{bahdanau2014neural,luong2015effective} between two sequences in a multi-branch fashion \cite{szegedy2015going}. Within the encoder or the decoder, each layer first computes attention between two copies of the same sequence (self-attention). In the decoder, this step is followed by an attention over the encoder output sequence. The last step in each layer is a two-layered time-distributed feed-forward network, with a hidden size larger than its input and output. 
Attention and feed-forward layers are characterized by a position-invariant processing of their input. Thus, in order to enrich input embeddings in source and target with positional information, they are summed with positional vectors of the same dimension $d$, which are computed with the following trigonometric encoding ($\text{PE}$):

\begin{equation}
    \text{PE}(\text{pos}, 2i) = \sin\left(\frac{pos}{10000^\frac{2i}{d}}\right)
\end{equation}
\begin{equation}
    \text{PE}(\text{pos}, 2i+1) = \cos\left(\frac{pos}{10000^\frac{2i+1}{d}}\right) 
\end{equation}
\noindent 
for $i=1,\ldots,d/2$.

\subsection{Length encoding in summarization}
Recently,  an extension of the positional encoding \cite{takase2019} was proposed to model the output length for text summarization. The goal is achieved by computing the distance from every position to the end of the sentence. The new \textit{length encoding} is present only in the decoder network as an additional vector summed to the input embedding. The authors proposed two different variants. The first variant replaces the variable \textit{pos} in equations (1-2) with the difference 
$len - pos$, where \textit{len} is the sentence length. The second variant \emph{attempts} to model the proportion of the sentence that has been covered at a given position by replacing the constant 10000 in the denominator of equations (1-2) with $len$.\footnote{Notice that the denominator varies with $i$ according to a power function.} As decoding is performed at the character level, \textit{len} and \textit{pos} are given in number of characters. At training time, \textit{len} is the observed length of the reference summary, while at inference time it is the desired length.

\section{Methods}
We propose two methods to control the output length in NMT. In the first method we partition the training set in three groups according to the observed length ratio of the reference over the source text. The idea is to let the model learn translation variants by observing them jointly with an extra input token. The second method extends the Transformer positional encoding to give information about the remaining sentence length. With this second method the network can leverage fine-grained information about the sentence length. 

\subsection{Length Token Method}
Our first approach to control the length is inspired by \textit{target forcing} in multilingual NMT \cite{johnson2016google, ha2016toward}. We first split the training sentence pairs into three groups according to the target/source length ratio (in terms of characters). Ideally, we want a group where the target is shorter than the source (\textit{short}), one where they are equally-sized (\textit{normal}) and a last group where the target is longer than the source (\textit{long}). In practice, we select two thresholds $t_\text{min}$ and $t_\text{max}$ according to the length ratio distribution. All the sentence pairs with length ratio between $t_\text{min}$ and $t_\text{max}$ are in the \textit{normal} group, the ones with ratio below $t_\text{min}$ in \textit{short} and the remaining in \textit{long}. At training time we prepend a \texttt{length token} to each source sentence according to its group ($<$short$>$, $<$normal$>$, or $<$long$>$), in order to let a single network to discriminate between the groups (see Figure \ref{fig:token}). At inference time, the length token is used to bias the network to generate a translation that belongs to the desired length group.

\begin{figure*}[t]
    \centering
    \includegraphics[width=\textwidth]{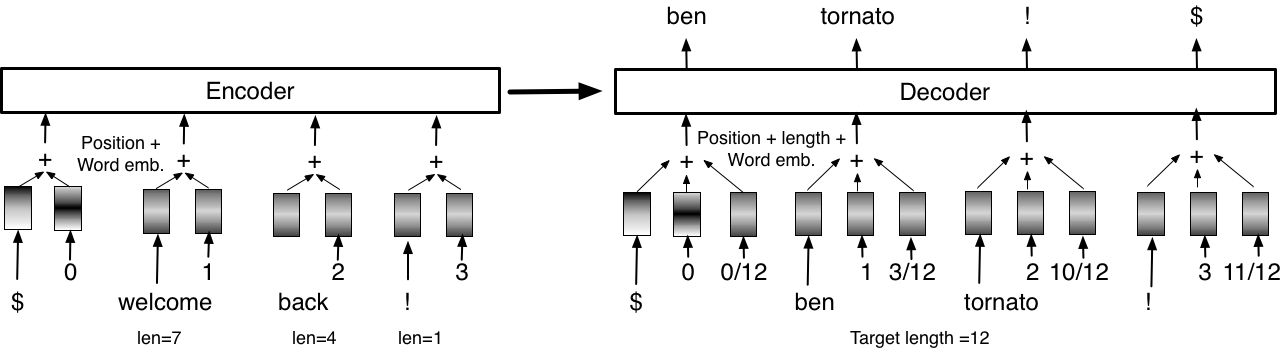}
    \caption{Transformer architecture with decoder input enriched with (relative) length embedding computed according to the desired target string length (12 characters in the example).}
    \label{fig:encoding}
\end{figure*}

\subsection{Length Encoding Method}\label{subsec:len-enc}
Inspired by \cite{takase2019}, we use length encoding to provide the network with information about the remaining sentence length during decoding. 
We propose two types of length encoding: \textit{absolute} and \textit{relative}. Let {\it pos} and {\it len} be, respectively, a token position and the end of the sequence, 
both expressed in terms of number characters. Then, the absolute approach encodes the remaining length: \begin{equation}
    \text{LE}_\text{abs}(\text{len}, \text{pos}, 2i) = \sin\left(\frac{len - pos}{10000^\frac{2i}{d}}\right)
\end{equation}
\begin{equation}
    \text{LE}_\text{abs}(\text{len}, \text{pos}, 2i+1) = \cos\left(\frac{len - pos}{10000^\frac{2i+1}{d}}\right)
\end{equation}
\noindent
where $i=1,\ldots,d/2$.

Similarly, the relative difference encodes the relative position to the end. This representation is made consistent with the absolute encoding by quantizing the space of the relative positions into a finite set of $N$ integers:
\begin{equation}
    \text{LE}_\text{rel}(\text{len}, \text{pos}, 2i) = \sin\left(\frac{q_N(pos / len)}{10000^\frac{2i}{d}}\right)
\end{equation}
\begin{equation}
    \text{LE}_\text{rel}(\text{len}, \text{pos}, 2i+1) = \cos\left(\frac{q_N(pos /len)}{10000^\frac{2i+1}{d}}\right)
\end{equation}

where $q_N: [0, 1] \rightarrow \{0, 1, .., N\}$ is simply defined as $q_N(x) = \lfloor{x \times N}\rfloor$. As we are interested in the character length of the target sequence, \textit{len} and \textit{pos} are given in terms of characters, but we represent the sequence as a sequence of BPE-segmented subwords \cite{sennrich2015neural}.
To solve the ambiguity, \textit{len} is the character length of the sequence, while \textit{pos} is the character count of all the preceding tokens. We prefer a representation based on BPE, unlike \cite{takase2019}, as it leads to better translations with less training time \cite{kreutzer2018learning,cherry2018revisiting}. During training, \textit{len} is the observed length of the target sentence, while at inference time it is the length of the source sentence, as it is the length that we aim to match. The process is exemplified in Figure \ref{fig:encoding}.

\subsection{Combining the two methods}
We further propose to use the two methods together to combine their strengths. In fact, while the length token acts as a soft constraint to bias NMT to produce short or long translation with respect to the source, actually no length information is given to the network.  On the other side, length encoding leverages information about the target length, but it is agnostic of the source length. 

\subsection{Fine-Tuning for length control}\label{ssec:fine-tuning}
Training an NMT model from scratch is a compute intensive and time consuming task. Alternatively, fine-tuning a pre-trained network shows to improve performance in several NMT scenarios \cite{zoph2016transfer, farajian2017multi, chu2018survey, luong2015stanford,thompson2018freezing}. For our length control approaches, we further propose to use fine-tuning an NMT model with length information, instead of training it from scratch.
By adopting a fine-tuning strategy, we specifically aim; {\it i}) to decouple the performance of the baseline NMT model from that of the additional length information, {\it ii}) control the level of aggressiveness that can come from the data (length token) and the model (length encoding), and {\it iii}) make the approaches versatile to any pre-trained model.
More importantly, it will allow to transform any NMT model to an output length aware version, while getting better improvements on the quality of the generated sequences.

\section{Experiments}
\subsection{Data and Settings}
Our experiments are run using the English$\rightarrow${Italian/German} portions of the \mbox{MuST-C} corpus \cite{mustc19}, which is extracted from TED talks, using the same train/validation/test split as provided with the corpus (see Table~\ref{tab:dataset}). As additional data, we use a mix of public and proprietary data for about $16$ million sentence pairs for English-Italian (En-It) and $4.4$ million WMT14 sentence pairs for the English-German (En-De).
While our main goal is to verify our hypotheses on a large data condition, thus the need to include proprietary data, for the sake of reproducibility in both languages we also provide results with systems only trained on TED Talks (small data condition). When training on large scale data we use Transformer with layer size of $1024$, hidden size of $4096$ on feed forward layers, $16$ heads in the multi-head attention, and $6$ layers in both encoder and decoder. When training only on TED talks, we set layer size of $512$, hidden size of $2048$ for the feed forward layers, multi-head attention with $8$ heads and again $6$ layers in both encoder and decoder. 

In all the experiments, we use the Adam \cite{kingma2014adam} optimizer with an initial learning rate of $1\times10^{-7}$ that increases linearly up to $0.001$ for $4000$ warm-up steps, and decreases afterwards with the inverse square root of the training step. The dropout is set to $0.3$ in all layers but the attention, where it is $0.1$. The models are trained with label smoothed cross-entropy with a smoothing factor of $0.1$. Training is performed on $8$ Nvidia V100 GPUs, with batches of $4500$ tokens per GPU. Gradients are accumulated for $16$ batches in each GPU \cite{ott2018scaling}. We select the models for evaluation by applying early stopping based on the validation loss.
All texts are tokenized with scripts from the Moses toolkit \cite{koehn2007moses}, and then words are segmented with BPE \cite{sennrich2015neural} with 32K joint merge rules.

\begin{table}[t!]
\small
\begin{center}
\begin{tabular}{l|rrr}
\bf Pairs    & \bf Train   & \bf Dev   & \bf Test \\ \hline
En-It (MuST-C)      & 241,618      & 1,210      & 2,574     \\
En-De (MuST-C)      & 229,703      & 1,423      & 2,641       \\
En-De (WMT14)       & 4,471,497     & 6,003      & 3,003        \\ 
\end{tabular}
\end{center}
\caption{\label{tab:dataset} Train, validation and test data size in number of examples.} \end{table}

For evaluation we take the best performing checkpoint on the dev set according to the loss.
The size of the data clusters used for the length token method and their corresponding target-source length ratios are reported in Table~\ref{tab:tokens}. 
The value of $N$ of the relative encoding is set to a small value (5), as in preliminary experiments we observed that a high value (100) produces results similar to the absolute encoding.

\begin{table}[t!]
\small
\begin{center}
\begin{tabular}{lr|rrrr}
        \bf Pairs        & Set       & \bf short         & \bf normal    & \bf long  & \bf Total \\ \hline
            En-It       & train    & 64185	    & 117589	        & 59844         & 241618 \\
                        & dev      & 247        & 576               & 487           & 1210   \\
                        & test     & 599        & 1200              & 775           & 2574   \\             En-De       & train    & 53417	    & 103951            & 72335         & 229703 \\ 
                        & dev      & 311        & 624               & 488           & 1423    \\
                        & test     & 554        & 1240              & 847           & 2641     \\ 
\hline
\hline
\multicolumn{2}{r|}{Length ratio}  &  $[0,1]$  & $(1,1.2]$ & $(1.2,\infty)$ & \\   
\end{tabular}
\end{center}
\caption{\label{tab:tokens} Train data category after assigning the length tokens (normal, short and long).}
\end{table}

\subsection{Models}
We evaluate our {\tt Baseline} Transformer using two decoding strategies:
{\it i}) a standard beam search inference (standard), and {\it ii}) beam search
with length penalty (penalty) set to $0.5$ to favor shorter translations \cite{wu2016google}. 

Length token models are evaluated with three strategies that correspond to the tokens prepended to the source test set at a time (short, normal, and long), and reported as {\tt Len-Tok}. Length encoding ({\tt Len-Enc}) models are evaluated in a length matching condition, i.e. output length has to match input length. We report the relative ({\tt Rel}) and absolute ({\tt Abs}) strategies of the approach as discussed in Section \ref{subsec:len-enc}. 
In the small data condition, we additionally evaluated how the fine-tuning strategy compares with a model trained from scratch. In the large data condition, we added a setting that combines both the length-token and length-encoding strategies.

\subsection{Evaluation}
To evaluate all models' performance we compute BLEU~\cite{papineni2002bleu} with the \emph{multi-bleu.perl} implementation\footnote{A script from the Moses SMT toolkit: http://www.statmt.org/moses} on the single-reference test sets of the En-It and En-De pairs. Given the absence of multiple references covering different length ratios, we also report n-gram precision scores (BLEU$^*$), by multiplying the BLEU score by the inverse of the brevity penalty \cite{papineni2002bleu}. BLEU$^*$ scores is meant to measure to what extent shorter translations are subset of longer translations. 

The impact on translation lengths is evaluated with the mean sentence-level length ratios between
MT output and source (LR$^{src}$) and between MT output and reference (LR$^{ref}$). 
\begin{table*}[t!]
\centering
\small
\begin{tabular}{lr|rrrr|rrrr}
\multicolumn{10}{c}{\textbf{Small Data}}                                                                                             \\\hline 
\bf Pairs           &               & \multicolumn{4}{c}{English-Italian}           &   \multicolumn{4}{c}{English-German}        \\\hline 
\bf Models          & \bf Strategy  & BLEU          & BLEU$^{*}$ & LR$^{src}$   & LR$^{ref}$             & \bf BLEU  & BLEU$^{*}$     & LR$^{src}$ & LR$^{ref}$      \\  \hline   
{\tt Baseline}      & standard        & 32.33         & 32.33     & 1.05          & 1.03      & \it 31.32  & \it 31.41     & 1.11      & 0.98 \\ 
                    & penalty       & \it 32.45     & \it 32.45 & 1.04      & 1.02  & 30.80      & 31.36 & \it 1.09      & 0.97  \\ \hline
                    
\multicolumn{10}{c}{Training from scratch} \\
\hline
                    & normal          & \it 32.54     & 32.54     & 1.04          & 1.02      & \it 31.48 & \bf 31.76 & 1.12      & 1.00   \\
 {\tt Len-Tok}      & short         & 31.62         & \it 32.90 & \it 0.97      & 0.95  & 28.53     & 31.15     & \bf 1.02  & 0.90    \\
                    & long          & 31.16         & 31.16     & 1.10          & 1.08      & 30.31     & 30.31     & 1.22      & 1.09  \\ 
\hline                    
{\tt Len-Enc Rel}       & match         & \it 30.96     & \it 30.96 & 1.03          & 1.01  & 29.04     & \it 30.67 & 1.06      & 0.95 \\ 
{\tt Len-Enc Abs}       & match         & 30.26         & 30.26     & \it 1.01      & 1.04      & 27.60     & 29.58     & \it 1.02  & 0.91 \\ \hline

\multicolumn{10}{c}{ Fine-tuning the baseline model}  \\\hline
                    & normal          & 32.41         & 32.41     & 1.05          & 1.02      & \bf 31.64 & \bf 31.64 & 1.12      & 0.99   \\
{\tt Len-Tok}       & short         & \bf 32.67     & \bf 32.80 & \bf 1.01          & 0.99      & 30.12     & 31.34     & \it 1.07  & 0.94    \\     
                    & long          & 32.00         & 32.00     & 1.06          & 1.04      & 31.35     & 31.35     & 1.15      & 1.02    \\\hline
{\tt Len-Enc Rel}       & match         & \it 32.10     & \it 32.10 & 1.05          & 1.03      & \it 30.73 & 31.58     & 1.09      & 0.97\\ 
{\tt Len-Enc Abs}       & match         & 31.24         & 31.24     & \it 1.02      & 1.01  & 30.29     & \it 31.65 & \it 1.07  & 0.95  \\ 
\end{tabular}
\caption{\label{tab:result_small} Performance of the baseline and models with length information trained from scratch and or by fine-tuning, in terms of BLEU, BLEU$^*$, mean length ratio of the output against the source (LR$^{src}$) and the reference (LR$^{ref}$). 
{\it italics} shows the best performing model under each category, while {\bf bold} shows the wining strategy.}
 
\end{table*}

\section{Results} We performed experiments in two conditions: small data and larger data. In the small data condition we only use the MuST-C training set. 
In the large data condition, a baseline model is first trained on large data, then it is fine-tuned on the MuST-C training set using the proposed methods.  
Tables \ref{tab:result_small} and \ref{tab:result_large} lists the results for the small and large data conditions. For the two language directions they show BLEU and BLEU* scores, as well as the average length ratios. 

\subsection{Small Data condition}

The baselines generate translations longer than the source sentence side, with a length ratio of 1.05 for Italian and 1.11 for German. Decoding with length penalty (penalty) slightly decreases the length ratios but they are still far from our goal of LR$^{src}$=1.00.

\noindent
\textbf{Fine-tuning.} A comparison  of the models trained from scratch (central portion of Table \ref{tab:result_small}) with their counterparts fine-tuned from the baseline (last portion of Table \ref{tab:result_small}) shows that the models in the first group generally generate shorter translations, but of worse quality. Additionally, the results with fine-tuning are not much different from the baseline.
Existing models can be enhanced to produce shorter sentences, and little variation is observed in their translation quality.

\noindent
\textbf{Length tokens.} Fine-tuning with \texttt{Len-Tok} (Fourth section in Table \ref{tab:result_small}) gives a coarse-grained control over the length, while keeping BLEU scores similar to the baseline or slightly better. Decoding with the token normal leads to translations slightly shorter than the baseline for En-It (LR$^{src}$=1.05 and LR$^{ref}$=1.02), while the token small strongly reduces the translation lengths up to almost the source length (LR$^{src}$=1.01). In the opposite side, the token long generates longer translations which are slightly worse than the others (32.00). A similar behavior is observed for En-De, where the LR$^{src}$ goes from 1.12 to 1.07 when changing normal with short, and to 1.15 with long. The results with the token long are not interesting for our task and are given only for the sake of completeness.

\noindent
\textbf{Length Encoding.} The last section of Table \ref{tab:result_small} lists the results of using length encoding ({\tt Len-Enc}) relative (\texttt{Rel}) and absolute (\texttt{Abs}). The two encodings lead to different generated lengths, with \texttt{Abs} being always shorter than \texttt{Rel}. Unfortunately, these improvements in the lengths correspond to a significant degradation in translation quality, mostly due to truncated sentences.

\subsection{Large data condition}
\begin{table*}[t!]
\small
\centering
\begin{tabular}{lr|rrrr|rrrr}

\multicolumn{10}{c}{\textbf{Large Data Condition}} \\ \hline 
\bf Pairs                       &               & \multicolumn{4}{c|}{English-Italian}                       &   \multicolumn{4}{c}{English-German}        \\\hline 
\bf Models                      & \bf Strategy  & BLEU      & BLEU$^{*}$    & LR$^{src}$    & LR$^{ref}$    & \bf BLEU  & BLEU$^{*}$    & LR$^{src}$    & LR$^{ref}$      \\  \hline   
{\tt Baseline}                  & standard      & 35.46     & 35.46         & 1.05          & 1.03          & \it 33.96     & 34.06         & 1.13          & 0.99  \\ 
                                & penalty       & \bf 35.75 & 35.75         & 1.04          & 1.01          & 33.64     & 34.19         & 1.11          & 0.98  \\ \hline

                                & normal        & \it 35.48     & 35.48           & 1.05          & 1.02          & \bf 34.10 & \bf 34.24     & 1.12          & 1.00    \\
{\tt Len-Tok}                   & short         & 35.39     & \bf 36.22     & \bf 1.00          & 0.98          & 30.61     & 33.27         & 1.05          & 0.93   \\
                                & long          & 34.71     & 34.71         & 1.08          & 1.05          & 33.46     & 33.46         & 1.21          & 1.08   
\\\hline
    
{\tt Len-Enc Rel}                   & match         & \it 35.18 & \it 35.18         & \bf 1.01          & 0.99          &  \it 33.61    &  \it 33.74        &  1.11         &  0.98 \\ 
{\tt Len-Enc Abs}                   & match         & 33.86     & 33.86         & 1.02          & 1.00          &  30.79    &  33.29        &  1.03         &  0.92  \\ \hline

{\tt Tok+Enc Rel}                   & short         & 34.51   &  35.91          & 0.96      & 0.94      & 30.08   &  32.62          &  \bf 1.01     &  0.90 \\
                                & normal          & \it 35.40   & \it 35.40          &  \bf 1.02         & 0.99          &  \it 33.41   &  \it 34.09         &  1.08         &  0.96  \\
{\tt Tok+Enc Abs}                   & short         & 33.96   &  33.96          &  \bf 1.01          & 0.99          &  29.28    &  32.28        &  \bf 1.01        &  0.90  \\
                                & normal          & 33.90    &  33.90       &  \bf 1.01          & 1.00          &  31.19  &  33.61          &  1.03        &  0.92  \\

\end{tabular}
\caption{\label{tab:result_large} Large scale experiments comparing the baseline, length token, length encoding and their combination.}
\end{table*}
Our \texttt{Baseline}s for the large data condition generate sentences with length ratios over the source comparable to the small data condition (LR$^\text{src}$ and LR$^\text{ref}$), but with better translation quality: 35.46 BLEU points for En-It and 33.96 for En-De. Length penalty slightly reduces the length ratios, which results in a 0.3 BLEU points improvement in Italian and -0.3 in German because of the brevity penalty. In the latter case, the BLEU* is slightly better than the standard baseline output. Also for the large data condition, while the length penalty slightly helps to shorten the translations, its effect is minimal and insufficient for our goal. 

\noindent
\textbf{Length tokens.} In En-It there is no noticeable difference in translation quality between the tokens normal and short, while there is a degradation of $\sim0.7$ points when using long. This last result is consistent with the ones observed before. Also in this case the token short does not degrade the BLEU score, and obtains the highest precision BLEU* with 36.22. In En-De we obtain the best results with token normal (34.46), which matches the length distribution of the references. The token short generates much shorter outputs (LR$^\text{src}$=1.05), which are also much shorter than the reference (LR$^\text{ref}=0.93$). Consequently the BLEU score degrades significantly (30.61), and also the BLEU* is 1 point lower than with the token normal. Longer translations can be generated with the token long, but they always come at the expense of lower quality.

\noindent
\textbf{Length encoding.} For En-It, \texttt{Len-Enc Rel} in Table \ref{tab:result_large} achieves a LR$^\text{src}$ of 1.01 with a slight degradation of $0.3$ BLEU points over the baseline, while in the case of \texttt{Abs} the degradation is higher (-1.6) and LR$^\text{src}$ is similar (1.02). Also in En-De the degradation of \texttt{Rel} over the baseline is only -0.3, but the reduction in terms of LR$^\text{src}$ is very small (1.11 vs 1.13). On the other side, \texttt{Abs} produces much shorter translations (1.03 LR$^\text{src}$) at the expense of a significantly lower BLEU score (30.79). When computing the BLEU* score, the absolute encoding is only 0.45 points lower than the relative encoding (33.29 vs 33.74), but -0.8 lower than the baseline.

\noindent
\textbf{Token + Encoding.} So far, we have observed generally good results using the token method and translating with the tokens short and normal. while the length encoding generally produces a more predictable output length, in particular for the absolute variant. In the last experiment, we combine the two methods in order to have a system that can capture different styles (short, normal, long), as well as explicitly leveraging length information. The results listed in the last portion of Table \ref{tab:result_large} (\texttt{Tok+Enc}) show that the relative encoding \texttt{Rel} produces better translations than \texttt{Abs}, but again it has less predictability in output length. For instance, in En-It the LR$^\text{src}$ of \texttt{Rel} is 0.96 with token short and 1.02 with normal, while for En-De it is 1.01 with short and 1.08 with normal. On the other side, the \texttt{Abs} produces LR$^\text{src}$ of 1.01 with both tokens in En-It and also with short in En-De, and it increases to only 1.03 with normal. 

\noindent
\textbf{Controlling output length.} In order to achieve LR$^\text{src}$ as close as possible to 1.0, we set the target length during generation equal to the source length when using the length encoding methods. However, one advantage of length encoding is the possibility to set the target length to modify the average output length. We illustrate this option by using the \texttt{Tok+Enc Rel} system for En-It, and translating with the tokens normal or short and different scaling factors for the target length. The results, listed in Table \ref{tab:result_decoding}, show that we are able to approach an LR$^{src}$ of 1.0 with both tokens and the BLEU score is not affected with token normal (35.45) or improves with token short (35.11). 

\noindent
\textbf{Discussion.} Length token is an effective approach to generate translations of different lengths, but it does not allow a fine-grained control of the output lengths and its results depend on the partition of the training set into groups, which is a manual process. Length encoding allows to change the output length, but the two variants have different effects. Absolute encoding is more accurate but generates sentences with missing information. The relative encoding produces better translations than the absolute encoding, but its control over the translation length is more loose.
The increased length stability is captured by the standard deviation of the length ratio with the source, which is $0.14$ for length tokens, $\sim0.11$ for relative encoding and $\sim0.07$ for absolute encoding.
The advantage of the combined approach is that it can generate sentences with different style to fit different length groups, and the output length can also be tuned by modifying the target length, while no important quality degradation is observed. Additionally, the standard deviation of the lengths is the same as for the length encoding used.

\begin{table}[t]
\small
\centering
\begin{tabular}{lr|rrrr}
\bf Token     & \bf Scale  & BLEU      & BLEU$^{*}$    & LR$^{src}$    & LR$^{ref}$ \\\hline
        &   1.00    &  34.51     &  \bf 35.91     & 0.96  & 0.94      \\
short   &   1.10    &  34.82     &  35.60     & 0.98  & 0.96      \\
        &   1.20    &  \it 35.11     &  35.25     & \bf 0.99  & 0.97      \\\hline
        &   1.00    &  \bf 35.40     &  35.40     & 1.02  & 0.99       \\
normal  &   0.98    &  \bf 35.49     &  35.49     & 1.01  & 0.99       \\
        &   0.93    &  \bf 35.46     &  \it 35.67     & \bf 1.00  & 0.98       \\
\end{tabular}
\caption{Results for En-It with \texttt{Tok+Enc Rel} by scaling the target length with different constant factors.}
\label{tab:result_decoding}
\end{table}

\subsection{Human Evaluation and Analysis}\label{subsec:human-eval} After manually inspecting the outputs of the best performing models under the large data 
condition, we decided to run a human evaluation only for the En-It {\tt Len-Tok} model.  
As our ultimate goal is to be able to generate shorter 
translations and as close as possible to the length of the source sentences, we focused
the manual evaluation on the Short output class and aimed to verify possible losses
in quality with respect to the baseline system.  We ran a head-to-head evaluation 
\footnote{We used crowd-sourcing via figure-eight.com.} on the first 10 sentences of each test talk, for a total of 270 sentences, by asking annotators to blindly rank the two system outputs (ties were also permitted) in terms of quality with respect to a reference translation.\footnote{Evaluators were asked to tell which version of the sentence was
best or if they were equivalent, given that a version is good if both the meaning of the reference is preserved and the grammar is correct.}
We collected three judgments for each output, from 19 annotators, for a total of 807 scores (one sentence had to be discarded). Inter-annotator agreement measured with Fleiss' kappa was 0.35 (= fair agreement).  Results reported in Table~\ref{tab:man-eval} confirm the small differences observed in BLEU scores: there are only a 4\% more wins for the Baseline and almost 60\% of ties. The small degradation in quality of the 
shorter translations is statistically significant\footnote{We used randomization tests with 15K repetitions  \cite{noreen1989}.} ($p<0.05$), as well as their difference in length  ($p<0.001$).

\begin{table}[t]
\begin{tabular}{l|rr}
                          & \% of Wins & LR$^{src}$ \\                       \hline
Baseline                  & 21.96  & 1.06\\
Len-Tok             & 17.99  & 1.01\\
P-value                   & $<$ 0.05 & $<$ 0.001\\
\end{tabular}
\caption{Manual evaluation on En-It (large data) ranking translation quality of the baseline ({\tt standard}) and token short translation 
against the reference translation.}
\label{tab:man-eval}
\end{table}

Notice that the evaluation was quite severe towards the shorter translations, as even small changes of the meaning could affect the ranking.   After the manual evaluation, we analyzed sentences in which shorter translations were unanimously judged equal or better than the standard translations. We hence tried to identify the linguistic skills involved in the generation of shorter translations, namely: (i) use of abbreviations, (ii) preference of simple verb tenses over compound tenses, (iii) avoidance of not relevant adjective, adverbs, pronouns and articles, (iv) use of paraphrases. Table~\ref{tab:examples2} shows examples of the application of the above strategies as found in the test set.

\begin{table}[t]
\small
\begin{tabular}{l|l}
EN  & And we in the West couldn't understand \\
MT  & \rephrase{E noi occidentali} non riuscivamo a capire \\
MT* & \rephrase{In occidente} non riuscivamo a capire \\\\
\hline
EN & how much this would restrict freedom of speech\\
MT  & quanto \drop{questo} \tense{avrebbe limitato} la libert\`a\\
MT* & quanto \tense{limitasse} la libert\`a \\\\
\hline
EN & this is a really extraordinary honor for me\\
MT  & \drop{questo} \`e un onore \drop{davvero} straordinario per me\\
MT* & per me \`e un onore straordinario\\\\
\hline
EN  & And this was done  \\ 
MT  & E questo \tense{\`e stato} fatto in modo che\\
MT* & E questo \tense{fu} fatto in modo che \\

\end{tabular}
\caption{Examples of shorter translation fragments obtained by paraphrasing (italics), drop of words (red), and change of verb tense (underline).}
\label{tab:examples2}
\end{table}

\section{Related works}
As an integration of Section~2, we try to provide a more complete picture on previous work with 
seq-to-seq models to control the output length for text summarization, and on the use of tokens
to bias in different ways the output of NMT. 

In text summarization, \cite{kikuchi2016controlling} proposed methods to control output length 
either by modifying the search process or the seq-to-seq model itself, showing that the latter 
being more promising. \cite{fan2017controllable} addressed the problem similarly to our token 
approach, by training the model on data bins of homogeneous output length and conditioning the 
output on a length token. They reported better performance than \cite{kikuchi2016controlling}. 
Finally, \cite{takase2019} proposed the extension of the positional encoding of the transformer 
(cf. Section~2), reporting better performance than \cite{kikuchi2016controlling} and \cite{fan2017controllable}.

The use of tokens to condition the output of NMT started with the multilingual models  \cite{johnson2016google,ha2016toward}, and was then further applied to control the use
of the politeness form  in English-German NMT \cite{sennrich2016controlling}, in the 
translation from English into different varieties of the same language \cite{lakew2018neural}, 
for personalizing NMT to user gender and vocabulary \cite{michel2018extreme}, and finally
to perform NMT across different translation styles~\cite{niu2018multi}.

\section{Conclusion}
In this paper, we have proposed two solutions for the problem of controlling the output length of NMT. A first approach, inspired by multilingual NMT, allows a coarse-grained control over the length and no degradation in translation quality. A second approach, inspired by positional encoding, enables a fine-grained control with only a small error in the token count, but at the cost of a lower translation quality. A manual evaluation confirms the translation quality observed with BLEU score. In future work, we plan to design more flexible and context-aware evaluations which allow us to account for short translations that 
are not equivalent to the original but at the same time do not affect the overall meaning of the discourse.

%\newpage
\bibliography{biblio}

\begin{thebibliography}{10}
\providecommand{\url}[1]{#1}
\csname url@samestyle\endcsname
\providecommand{\newblock}{\relax}
\providecommand{\bibinfo}[2]{#2}
\providecommand{\BIBentrySTDinterwordspacing}{\spaceskip=0pt\relax}
\providecommand{\BIBentryALTinterwordstretchfactor}{4}
\providecommand{\BIBentryALTinterwordspacing}{\spaceskip=\fontdimen2\font plus
\BIBentryALTinterwordstretchfactor\fontdimen3\font minus
  \fontdimen4\font\relax}
\providecommand{\BIBforeignlanguage}[2]{{%
\expandafter\ifx\csname l@#1\endcsname\relax
\typeout{** WARNING: IEEEtran.bst: No hyphenation pattern has been}%
\typeout{** loaded for the language `#1'. Using the pattern for}%
\typeout{** the default language instead.}%
\else
\language=\csname l@#1\endcsname
\fi
#2}}
\providecommand{\BIBdecl}{\relax}
\BIBdecl

\bibitem{bahdanau2014neural}
D.~Bahdanau, K.~Cho, and Y.~Bengio, ``Neural machine translation by jointly
  learning to align and translate,'' \emph{arXiv preprint arXiv:1409.0473},
  2014.

\bibitem{sutskever2014sequence}
I.~Sutskever, O.~Vinyals, and Q.~V. Le, ``Sequence to sequence learning with
  neural networks,'' in \emph{Advances in neural information processing
  systems}, 2014, pp. 3104--3112.

\bibitem{bentivogli:CSL2018}
L.~Bentivogli, A.~Bisazza, M.~Cettolo, and M.~Federico, ``Neural versus
  phrase-based mt quality: An in-depth analysis on english-german and
  english-french,'' \emph{Computer Speech \& Language}, vol.~49, pp. 52--70,
  2018.

\bibitem{hassan18}
\BIBentryALTinterwordspacing
H.~Hassan, A.~Aue, C.~Chen, V.~Chowdhary, J.~Clark, C.~Federmann, X.~Huang,
  M.~Junczys{-}Dowmunt, W.~Lewis, M.~Li, S.~Liu, T.~Liu, R.~Luo, A.~Menezes,
  T.~Qin, F.~Seide, X.~Tan, F.~Tian, L.~Wu, S.~Wu, Y.~Xia, D.~Zhang, Z.~Zhang,
  and M.~Zhou, ``Achieving human parity on automatic chinese to english news
  translation,'' \emph{CoRR}, vol. abs/1803.05567, 2018. [Online]. Available:
  \url{http://arxiv.org/abs/1803.05567}
\BIBentrySTDinterwordspacing

\bibitem{laubi18}
\BIBentryALTinterwordspacing
S.~L{\"{a}}ubli, R.~Sennrich, and M.~Volk, ``Has machine translation achieved
  human parity? {A} case for document-level evaluation,'' in \emph{Proceedings
  of the 2018 Conference on Empirical Methods in Natural Language Processing},
  Brussels, Belgium, 2018, pp. 4791--4796. [Online]. Available:
  \url{https://aclanthology.info/papers/D18-1512/d18-1512}
\BIBentrySTDinterwordspacing

\bibitem{murray2018correcting}
K.~Murray and D.~Chiang, ``Correcting length bias in neural machine
  translation,'' in \emph{Proceedings of the Third Conference on Machine
  Translation: Research Papers}, 2018, pp. 212--223.

\bibitem{shi2016neural}
X.~Shi, K.~Knight, and D.~Yuret, ``Why neural translations are the right
  length,'' in \emph{Proceedings of the 2016 Conference on Empirical Methods in
  Natural Language Processing}, 2016, pp. 2278--2282.

\bibitem{rush2015neural}
A.~M. Rush, S.~Chopra, and J.~Weston, ``A neural attention model for
  abstractive sentence summarization,'' \emph{arXiv preprint arXiv:1509.00685},
  2015.

\bibitem{kikuchi2016controlling}
Y.~Kikuchi, G.~Neubig, R.~Sasano, H.~Takamura, and M.~Okumura, ``Controlling
  output length in neural encoder-decoders,'' \emph{arXiv preprint
  arXiv:1609.09552}, 2016.

\bibitem{fan2017controllable}
A.~Fan, D.~Grangier, and M.~Auli, ``Controllable abstractive summarization,''
  \emph{arXiv preprint arXiv:1711.05217}, 2017.

\bibitem{liu2018controlling}
Y.~Liu, Z.~Luo, and K.~Zhu, ``Controlling length in abstractive summarization
  using a convolutional neural network,'' in \emph{Proceedings of the 2018
  Conference on Empirical Methods in Natural Language Processing}, 2018, pp.
  4110--4119.

\bibitem{takase2019}
\BIBentryALTinterwordspacing
T.~Sho and O.~Naoaki, ``Positional encoding to control output sequence
  length,'' in \emph{Proceedings of the HLT-NAACL 2019}, 2019. [Online].
  Available: \url{http://arxiv.org/abs/1904.07418}
\BIBentrySTDinterwordspacing

\bibitem{vaswani2017attention}
A.~Vaswani, N.~Shazeer, N.~Parmar, J.~Uszkoreit, L.~Jones, A.~N. Gomez,
  {\L}.~Kaiser, and I.~Polosukhin, ``Attention is all you need,'' in
  \emph{Advances in Neural Information Processing Systems}, 2017, pp.
  6000--6010.

\bibitem{luong2015effective}
M.-T. Luong, H.~Pham, and C.~D. Manning, ``Effective approaches to
  attention-based neural machine translation,'' \emph{arXiv preprint
  arXiv:1508.04025}, 2015.

\bibitem{szegedy2015going}
C.~Szegedy, W.~Liu, Y.~Jia, P.~Sermanet, S.~Reed, D.~Anguelov, D.~Erhan,
  V.~Vanhoucke, and A.~Rabinovich, ``Going deeper with convolutions,'' in
  \emph{Proceedings of the IEEE conference on computer vision and pattern
  recognition}, 2015, pp. 1--9.

\bibitem{johnson2016google}
M.~Johnson, M.~Schuster, Q.~V. Le, M.~Krikun, Y.~Wu, Z.~Chen, N.~Thorat,
  F.~Vi{\'e}gas, M.~Wattenberg, G.~Corrado \emph{et~al.}, ``Google's
  multilingual neural machine translation system: Enabling zero-shot
  translation,'' \emph{arXiv preprint arXiv:1611.04558}, 2016.

\bibitem{ha2016toward}
T.-L. Ha, J.~Niehues, and A.~Waibel, ``Toward multilingual neural machine
  translation with universal encoder and decoder,'' \emph{arXiv preprint
  arXiv:1611.04798}, 2016.

\bibitem{sennrich2015neural}
R.~Sennrich, B.~Haddow, and A.~Birch, ``Neural machine translation of rare
  words with subword units,'' \emph{arXiv preprint arXiv:1508.07909}, 2015.

\bibitem{kreutzer2018learning}
J.~Kreutzer and A.~Sokolov, ``Learning to segment inputs for nmt favors
  character-level processing,'' 2018.

\bibitem{cherry2018revisiting}
C.~Cherry, G.~Foster, A.~Bapna, O.~Firat, and W.~Macherey, ``Revisiting
  character-based neural machine translation with capacity and compression,''
  in \emph{Proceedings of the 2018 Conference on Empirical Methods in Natural
  Language Processing}, 2018, pp. 4295--4305.

\bibitem{zoph2016transfer}
B.~Zoph, D.~Yuret, J.~May, and K.~Knight, ``Transfer learning for low-resource
  neural machine translation,'' \emph{arXiv preprint arXiv:1604.02201}, 2016.

\bibitem{farajian2017multi}
M.~A. Farajian, M.~Turchi, M.~Negri, and M.~Federico, ``Multi-domain neural
  machine translation through unsupervised adaptation,'' in \emph{Proceedings
  of the Second Conference on Machine Translation}, 2017, pp. 127--137.

\bibitem{chu2018survey}
C.~Chu and R.~Wang, ``A survey of domain adaptation for neural machine
  translation,'' \emph{arXiv preprint arXiv:1806.00258}, 2018.

\bibitem{luong2015stanford}
M.-T. Luong and C.~D. Manning, ``Stanford neural machine translation systems
  for spoken language domains,'' in \emph{Proceedings of the International
  Workshop on Spoken Language Translation}, 2015.

\bibitem{thompson2018freezing}
B.~Thompson, H.~Khayrallah, A.~Anastasopoulos, A.~D. McCarthy, K.~Duh,
  R.~Marvin, P.~McNamee, J.~Gwinnup, T.~Anderson, and P.~Koehn, ``Freezing
  subnetworks to analyze domain adaptation in neural machine translation,'' in
  \emph{Proceedings of the Third Conference on Machine Translation: Research
  Papers}, 2018, pp. 124--132.

\bibitem{mustc19}
M.~A. Di~Gangi, R.~Cattoni, L.~Bentivogli, M.~Negri, and M.~Turchi, ``{MuST-C:
  a Multilingual Speech Translation Corpus},'' in \emph{Proceedings of the 2019
  Conference of the North American Chapter of the Association for Computational
  Linguistics: Human Language Technologies, Volume 2 (Short Papers)},
  Minneapolis, MN, USA, June 2019.

\bibitem{kingma2014adam}
D.~Kingma and J.~Ba, ``Adam: A method for stochastic optimization,''
  \emph{arXiv preprint arXiv:1412.6980}, 2014.

\bibitem{ott2018scaling}
M.~Ott, S.~Edunov, D.~Grangier, and M.~Auli, ``Scaling neural machine
  translation,'' in \emph{Proceedings of the Third Conference on Machine
  Translation: Research Papers}, 2018, pp. 1--9.

\bibitem{koehn2007moses}
P.~Koehn, H.~Hoang, A.~Birch, C.~Callison-Burch \emph{et~al.}, ``Moses: Open
  source toolkit for statistical machine translation,'' in \emph{Proc. of ACL},
  2007.

\bibitem{wu2016google}
Y.~Wu, M.~Schuster, Z.~Chen, Q.~V. Le, M.~Norouzi, W.~Macherey, M.~Krikun,
  Y.~Cao, Q.~Gao, K.~Macherey \emph{et~al.}, ``Google's neural machine
  translation system: Bridging the gap between human and machine translation,''
  \emph{arXiv preprint arXiv:1609.08144}, 2016.

\bibitem{papineni2002bleu}
K.~Papineni, S.~Roukos, T.~Ward, and W.-J. Zhu, ``Bleu: a method for automatic
  evaluation of machine translation,'' in \emph{Proceedings of the 40th annual
  meeting on association for computational linguistics}.\hskip 1em plus 0.5em
  minus 0.4em\relax Association for Computational Linguistics, 2002, pp.
  311--318.

\bibitem{noreen1989}
E.~Noreen, \emph{Computer-{Intensive} {Methods} for {Testing} {Hypotheses}:
  {An} {Introduction}}.\hskip 1em plus 0.5em minus 0.4em\relax Wiley, 1989.

\bibitem{sennrich2016controlling}
R.~Sennrich, B.~Haddow, and A.~Birch, ``Controlling politeness in neural
  machine translation via side constraints,'' in \emph{Proceedings of the 2016
  Conference of the North American Chapter of the Association for Computational
  Linguistics: Human Language Technologies}, 2016, pp. 35--40.

\bibitem{lakew2018neural}
S.~M. Lakew, A.~Erofeeva, and M.~Federico, ``Neural machine translation into
  language varieties,'' \emph{arXiv preprint arXiv:1811.01064}, 2018.

\bibitem{michel2018extreme}
P.~Michel and G.~Neubig, ``Extreme adaptation for personalized neural machine
  translation,'' \emph{arXiv preprint arXiv:1805.01817}, 2018.

\bibitem{niu2018multi}
X.~Niu, S.~Rao, and M.~Carpuat, ``Multi-task neural models for translating
  between styles within and across languages,'' \emph{arXiv preprint
  arXiv:1806.04357}, 2018.

\end{thebibliography}
\bibliographystyle{IEEEtran}

\end{document}